\documentclass[letterpaper, 10 pt, conference]{ieeeconf}  

\IEEEoverridecommandlockouts                              

\usepackage[caption=false,font=footnotesize]{subfig}
\usepackage{graphicx} 
\usepackage{gensymb}

\usepackage{algorithm}
\usepackage{algorithmic}

\title{\LARGE \bf
Explicit-risk-aware Path Planning with Reward Maximization
}

\author{Xuesu Xiao$^{1}$, Jan Dufek$^{1}$, and Robin Murphy$^{1}$
\thanks{$^{1}$Xuesu Xiao, Jan Dufek, and Robin Murphy are with the Department of Computer Science and Engineering,
        Texas A\&M University, College Station, TX 77843
        {\tt\small \{xiaoxuesu, dufek, robin.r.murphy\}@tamu.edu}}%
}

\begin{document}

\maketitle
\thispagestyle{empty}
\pagestyle{empty}

\begin{abstract}


This paper develops a path planner that minimizes risk (e.g. motion execution) while maximizing accumulated reward (e.g., quality of sensor viewpoint) motivated by visual assistance or tracking scenarios in unstructured or confined environments. In these scenarios, the robot should maintain the best viewpoint as it moves to the goal. However, in unstructured or confined environments, some paths may increase the risk of collision; therefore there is a tradeoff between risk and reward. Conventional state-dependent risk or probabilistic uncertainty modeling do not consider path-level risk or is difficult to acquire. This risk-reward planner explicitly represents risk as a function of motion plans, i.e., paths. Without manual assignment of the negative impact to the planner caused by risk, this planner takes in a pre-established viewpoint quality map and plans target location and path leading to it simultaneously, in order to maximize overall reward along the entire path while minimizing risk. Exact and approximate algorithms are presented, whose solution is further demonstrated on a physical tethered aerial vehicle. Other than the visual assistance problem, the proposed framework also provides a new planning paradigm to address minimum-risk planning under dynamical risk and absence of substructure optimality and to balance the trade-off between reward and risk.

\end{abstract}

\section{INTRODUCTION}
Planning to achieve high mission performance while facing risk is a common trade-off in robotic motion or path planning. Such intelligent planning systems must be capable of deciding when to take risks to achieve high mission performance and when to be conservative due to the lack of reward. The usages of Unmanned Aerial Vehicles (UAVs) in situations such as Urban Search And Rescue (USAR), nuclear operations, disaster robotics \cite{murphy2014disaster}, etc., are examples where the execution of motion inherently entails taking risk. 

One particular example is the visual assistance problem, which is motivated by the Fukushima Daiichi nuclear decommissioning task. Teleoperation of a robot in remote confined and cluttered spaces with only first person view from robot's onboard camera is difficult due to perceptual limitations, such as lack of depth perception. Using a separate tele-operated robot can partially solve the problem by providing extra viewpoints, but also introduces problems, e.g., extra operators and teamwork demands. Therefore, autonomous visual assistants have been developed to provide better situational awareness in the remote field while avoiding extra cost of human labor and teamwork \cite{xiao2017visual, xiao2018motion}. 

However, flying autonomously in unstructured or confined environments for optimal visual assistance performance entails risks from different aspects. Performing motions to navigate to or maintain at a good viewpoint may be risky and put the safety of the agent at danger. This paper formulates the trade-off between reward and risk using a novel problem definition. In contrast to the traditional state-dependent or implicit probabilistic risk representation and Chance-Constrained (CC) process or Robust Model Predictive Control (RMPC), this paper proposes an explicit risk representation based on entire motion plan and a planner that maximizes the overall path utility value, defined as the ratio between total reward and risk. The planned path is implemented on a real robot and the physically collected reward and encountered risk are presented. 

The rest of the paper is organized as follows: Sec. \ref{sec::related_work} provides related work. Sec. \ref{sec::explicit_risk_representation} discusses the proposed explicit risk representation as a function of entire path. Sec. \ref{sec::approach} formally formulates the problem and proposes the exact and approximate algorithms. Sec. \ref{sec::experiments} presents physical demonstration of the algorithm on a real tethered aerial visual assistant robot. Sec. \ref{sec::conclusion} concludes the paper. 

\section{RELATED WORK}
\label{sec::related_work}
This section reviews current approaches to represent risk and to balance the trade-off between reward and risk. 
\subsection{Risk Representation}
In the literature, motion risk associated with navigating in unstructured of confined environments is either represented as (1) a risk function of the state or as (2) sensing and action uncertainty. 

\cite{soltani2004fuzzy} represented the workspace by three layers: distance, hazard data, and visibility layer. The risk related with each layer was a function of the particular state. \cite{de2011minimum} associated UAV flight risk at a certain location with this location's ground orography. \cite{vian1989trajectory, zabarankin2002optimal, gu2006comprehensive} adopted a similar approach and also assumed risk to be a function of location only. Even with data-driven approaches, researchers estimated potential risk of a certain state based on historical record, including ocean Automated Identification System (AIS) \cite{pereira2011toward, pereira2013risk} and traffic data \cite{krumm2017risk}. Assuming risk as a function of only state neglects those risk elements caused by the execution of an entire path \cite{murphy2014disaster, agarwal2014characteristics}, such as path tortuosity (number of turns needed to traverse path), which this research aims at including. 

Another important branch of risk representation is through implicit probabilistic models in belief space. Risk is trivial given a perfectly known world and action model. It is the model uncertainty that introduces risk into path execution, e.g., not knowing exactly where the robot is may lead to collision with obstacles. Belief Roadmap (BRM) \cite{prentice2010belief}, Rapidly-exploring Random Belief Trees \cite{bry2011rapidly}, linear-quadratic controller based on an ensemble of paths \cite{van2011lqg}, local optimization over Partially Observable Markov Decision Process (POMDP) \cite{van2012motion}, Feedback-based Information Roadmap (FIRM) \cite{agha2014firm} were used, representing risk as model uncertainty, to plan safe path. Most works were done in simulation with theoretical belief models. However, when planning with real robot in physical environments, a convincing method to quantify the probabilistic model is difficult to acquire.  

\subsection{Reward-risk Trade-off}
The trade-off between reward and risk was mostly addressed as chance or probability of success/failure. \cite{luders2010chance, luders2013robust} proposed chance-constrained rapidly-exploring random tree (CC-RRT) approaches, which used chance constraints to guarantee probabilistic feasibility at each time step and over entire trajectory. Another popular approach to handle reward and risk is to use (PO)MDP. As standard MDP inherently contains reward but not risk, researchers have looked into representing risk as negative reward (penalty) \cite{pereira2013risk} or constraints (C-POMDP) with unit cost for constraint violation \cite{undurti2010online, undurtifunction, undurti2011decentralized}. Going beyond unit cost, CC-POMDP was proposed by \cite{santana2016rao}, which was based on a bound on the probability (chance) of some event happening during policy execution. RMPC is another alternative, with an emphasis on risk allocation, i.e., to allocate more risk for more rewarding actions \cite{ono2008efficient, vitus2011feedback}. 

All existing methods require an artificial assignment of the adverse impact caused by risk, in the form of negative reward (penalty), unit cost, or bound on probability. Such assignment is not clear and a desired value may not even exist \cite{undurti2010online}. It is subjective to human bias and hard to determine, e.g., how to define the value of better situational awareness compared to the cost of losing the robot. This paper avoids the necessity of such artificial assignment, e.g., manually arbitrating the value of either maximum acceptable risk or minimum expected reward. A utility function of the ratio between reward and risk along the entire path is proposed, as a measurement of how much risk is taken to achieve one unit of reward. In addition, the desirable goal state is not specified to the planner beforehand, but discovered by the planner based on optimal utility during the planning process. 

\section{EXPLICIT RISK REPRESENTATION}
\label{sec::explicit_risk_representation}
This section presents the idea of explicit states risk and path risk and explains why and how risk is represented as a function of the entire path, not only individual or simple summation of individual states. Risk is explicitly represented in physical space. 

\begin{figure}[]
\centering
\includegraphics[width=1\columnwidth]{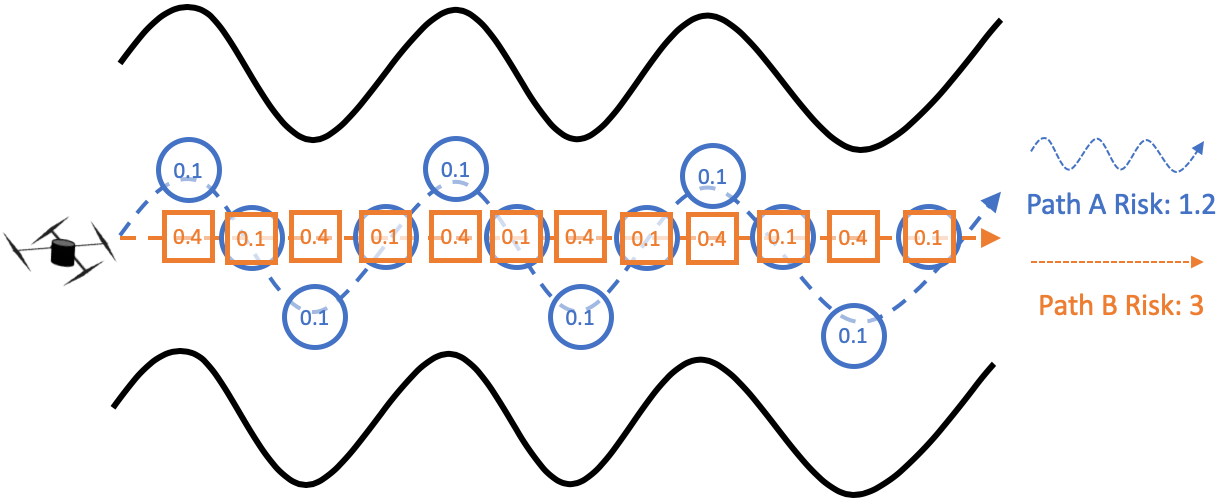}
\caption{State-Dependent Risk of Two Path Options: based on simple summation of state-dependent risk only, path B is much riskier than path A. However, extra tortuosity adds path-dependent risk to path A. }
\label{fig::two_path}
\end{figure}

The risk associated with executing a path in unstructured or confined spaces could be partially reflected by the risk of each individual state on the path. However, modeling risk only as a function of state ignores another important aspect of risk: path-dependent risk. Fig. \ref{fig::two_path} shows two possible paths for a UAV to traverse through a twisty and narrow corridor. Path A goes through a series of safe states (0.1 risk each) since the distance to closest obstacle is maximized along the entire path. However, taking six turns entails taking extra risk \cite{murphy2014disaster, agarwal2014characteristics}. Path B is safe in terms of a straight and easy path, but it has to go through risky states which are close to obstacles. Only considering safer states will prefer path A while only considering safer path will prefer path B. However, both states risk and path risk will cause challenges when executing the path. Therefore, both types of risk need to be considered in a comprehensive risk representation. 

States risk could be determined as a function of state only. It does not need any history information about how the robot comes to this state. Typical risk elements that will cause states risk are distance to closest obstacle, visibility (obstacle density), action length, access element, etc. The severity of those risk elements, defined as a numerical risk index, could be uniquely determined by the state alone. Risk elements that will cause path risk include tortuosity, path length, etc. Only knowing the state is not sufficient to determine the risk of coming to this state and historical path information is necessary for path-level risk. Using this general categorization, vehicle-specific risk elements could be added and fall into either states risk or path risk. UAV altitude could be defined for aerial vehicles and terrain stability for ground robot. For our particular tethered aerial visual assistant \cite{xiao2017visual}, tether length \cite{xiao2018indoor} and number of contact points \cite{xiao2018motion} are other examples of path risk elements.  

All these individual risk elements are mapped into a value between 0 and 1, indicating the unit risk level caused by each element using either membership function of fuzzy logic or normalization. All risk elements within each category (states risk and path risk) are summed up and further normalized between 0 and 1. For states risk, this value in [0, 1] represents the risk level of a certain state and needs to be integrated over time to reflect states risk along the entire path. For path risk, this value only needs to be evaluated once based on the path. The integrated states risk and standalone path risk will be combined to quantify the risk of executing that particular path using either fuzzy rules or weighted sum.

\section{RISK-AWARE REWARD-MAXIMIZING PATH PLANNING}
\label{sec::approach}
In this section, the visual assistance problem is formally defined as a graph search problem. It is shown that this problem is well-defined, i.e., there exists one, if not more, optimal solution to this problem.  It is believed that finding this optimal solution approximates finding the optimal visual assistance behavior. An exact algorithm is provided, which is guaranteed finding the optimal solution. The NP-hardness is further discussed. In order to solve the problem more efficiently, a two-stage approximate algorithm is proposed to balance the trade-off between risk and reward. 

\subsection{Problem Definition}
The planning space is defined to be an undirected graph converted from the 3-D occupancy grid map of the workspace. Each vertex of the graph corresponds to a viable state of the visual assistant. Based on a separate study on viewpoint quality, each vertex will be assigned a reward value, representing the visual assistance quality from that state. Due to the fact that visual assistance is about providing continuous visual feedback of the remote environment, the agent needs to maximize the reward along the entire path to maintain the operator's situational awareness in a continuous manner. So rewards are collected by visiting individual states and accumulated along the entire path. Motion risk is explicitly represented as a function of path. The agent's sensor and action models are assumed to be deterministic, given that the model uncertainties are already embedded in the risk representation. The agent starts at a given start and does not have a target. This is because manually arbitrating a target location based on reward only cannot consider risk properly. The best viewpoint may be very risky and it does not worth to take the risk to go there. The agent plans its actions to navigate the graphical state space with the goal to maximize path utility, the ratio between reward and risk. 

Given a finite graph and a start, there is a finite number of simple paths. Each of those path has a collective reward value by summing up the rewards of all visited vertices, and motion risk could be evaluated based on the path. So a utility value exists for every path. This means there exists one, if not more, path with maximum utility. So this problem is well-defined and the maximum utility path could be found, at least through brutal force approach that enumerates all simple paths. 

\subsection{Exact Algorithm}
Given a graph to represent the work space: $\mathcal{G} = (\mathcal{V}, \mathcal{E})$ with $\mathcal{V} = \{v_1, v_2, ..., v_n\}$ to be the vertex set, and $\mathcal{E} = \{e_1, e_2, ..., e_m\}$ to be all the edges connecting the vertices, $v_{start}$ represents start location. Reward map from the separate study is matched with $\mathcal{V}$ so that a reward value could be computed from a look-up table $rewards$ for any vertex $v_i$. Alg. \ref{alg::find_all_simple_path} shows the recursive Depth-First-Search (DFS) based algorithm to recursively find all simple paths: the main function calls Alg. \ref{alg::find_all_simple_path} and passes in $\mathcal{G}$, $v_{start}$, $path$ as one single vertex $v_{start}$, the $rewards$ look-up table, and $current\_reward$ as $0$. A discount factor $\gamma$ between $[0, 1]$ is used to determine how much current reward is favored over history rewards. When expanding from vertex $u$ to $v$, it recursively calls itself on vertex $v$ with updated information. 

\begin{algorithm}[]
 \caption{Evaluate\_All\_Simple\_Paths}
 \begin{algorithmic}[1]
 \renewcommand{\algorithmicrequire}{\textbf{Input:}}
 \renewcommand{\algorithmicensure}{\textbf{Global Variable:}}
 \REQUIRE $\mathcal{G}$, $u$, $path$, $rewards$, $current\_reward$, $\gamma$
 \ENSURE  $all\_simple\_paths$, $path\_utilities$
 \FOR {each edge $(u, v)\in\mathcal{G}$}
 	\IF {$v \notin path$}
 		\STATE $path\leftarrow path \cup v$
		\STATE $path\_risk\leftarrow evaluate(path)$
		\STATE $current\_reward \leftarrow \gamma*current\_reward+rewards(v)$
		\STATE $utility$ $\leftarrow$ $current\_reward$ / $path\_risk$
		\STATE $path\_utilities \leftarrow path\_utilities \cup utility$
		\STATE $all\_simple\_paths \leftarrow all\_simple\_paths \cup path$
		\STATE Evaluate\_All\_Simple\_Paths ($\mathcal{G}$, $v$, $path$, $rewards$, $current\_reward$, $\gamma$)
		\STATE $path \leftarrow path \setminus v$
		\STATE $path\_risk\leftarrow evaluate(path)$
		\STATE $current\_reward \leftarrow \frac{(current\_reward-rewards(v))}{\gamma}$
	\ENDIF
 \ENDFOR
 \end{algorithmic}
 \label{alg::find_all_simple_path}
 \end{algorithm}

\begin{figure}[]
\centering
\subfloat[Exact Solution]{\includegraphics[width=0.5\columnwidth]{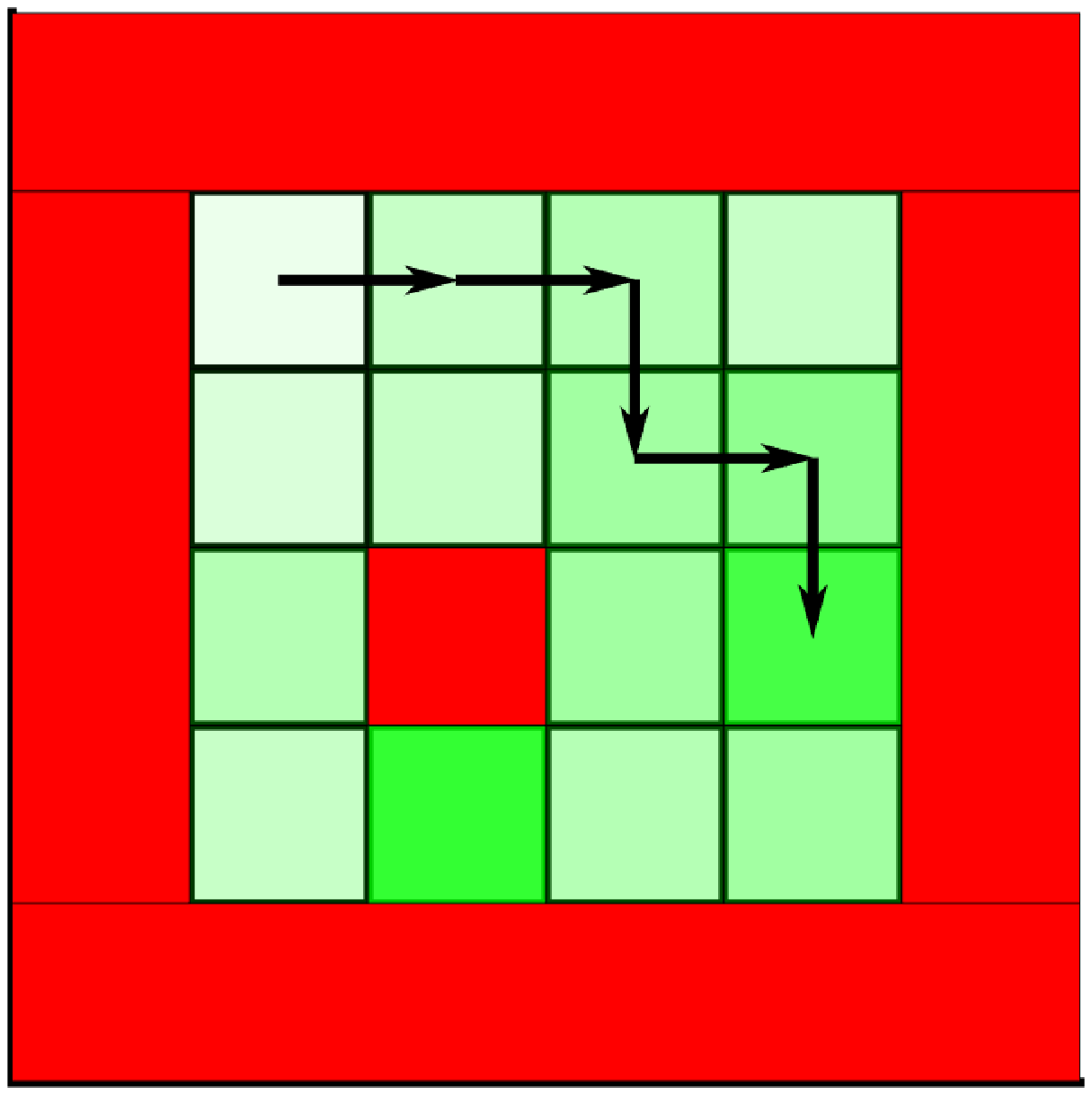}%
\label{fig::exact}}
\subfloat[Approximate Solution ]{\includegraphics[width=0.5\columnwidth]{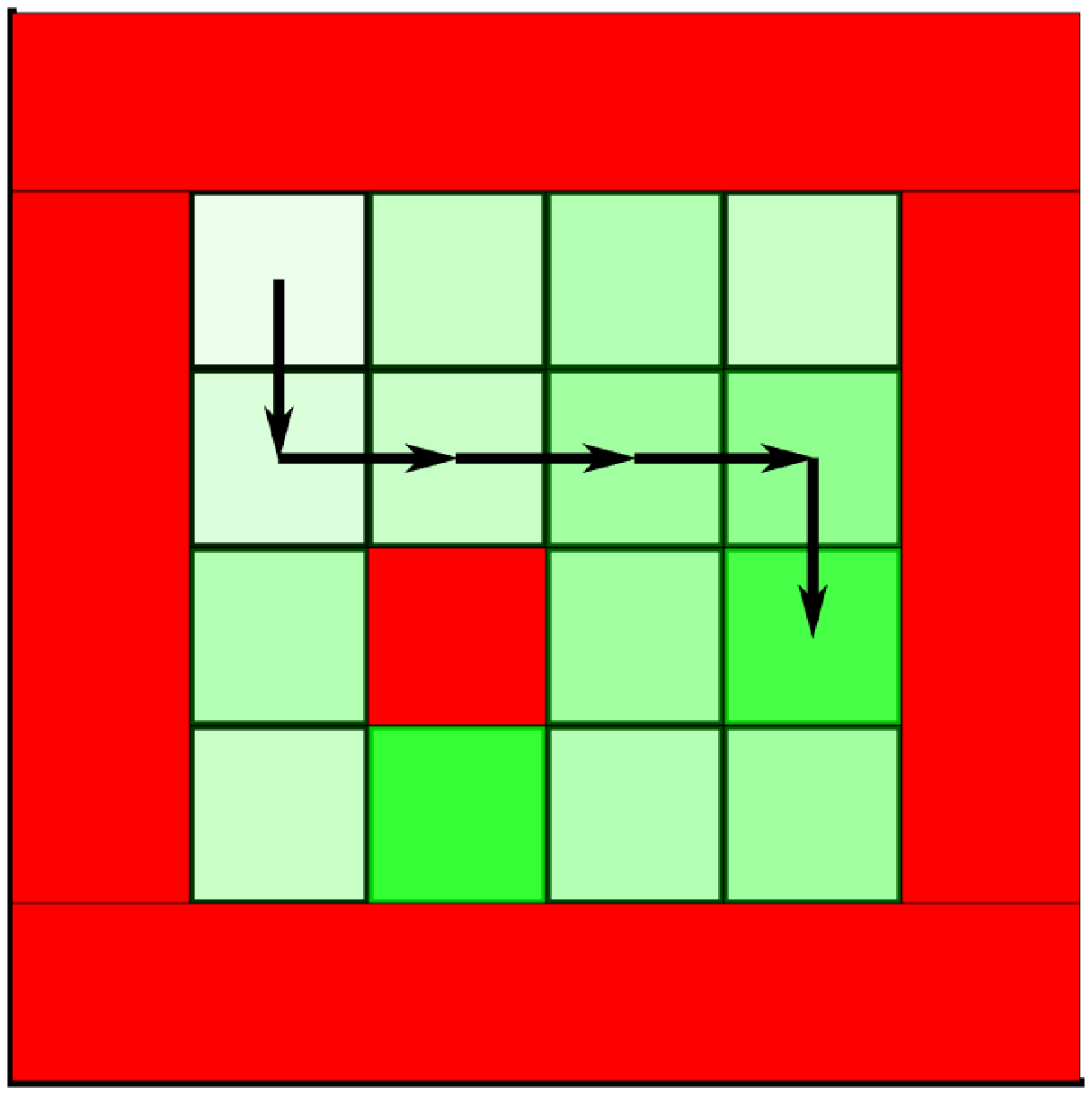}%
\label{fig::approximate}}
\caption{Exact and Approximate Algorithms on a Simple 4 by 4 Graph with 4 Connectivities: red cells denote obstacles and the greenness of each cell indicates reward value (viewpoint quality).}
\label{fig::both_algos}
\end{figure}

A simple example on a 2-D 4 by 4 occupancy grid with 4 connectivity is shown in Fig. \ref{fig::exact}. By enumerating all simple paths in this graph, the optimal utility path is found. For a small graph with only 16 vertices, however, there already exist 440 paths. The number of simple path exponentially increases to more than ten thousands even for a 5 by 5 graph. This is why exact brutal force algorithm is not practical for reasonable sized graphs. 

\subsection{NP-hardness}
\begin{figure}[]
\centering
\includegraphics[width=0.8\columnwidth]{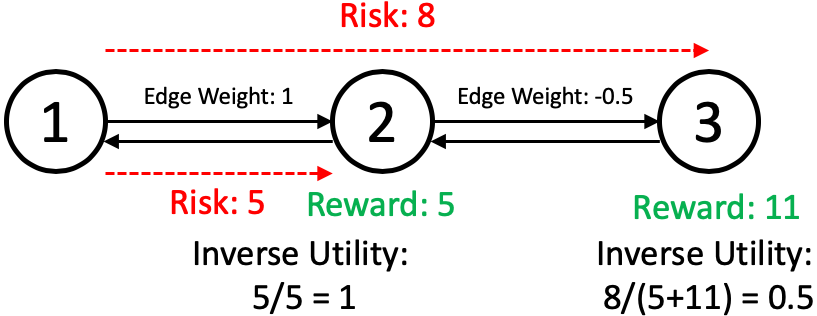}
\caption{Easy Conversion between Risk Reward (Utility) and Edge Weight Representation: going to vertex 2 has risk 5 and reward 5, which gives inverse utility 1. Going to vertex 3 via 2 has risk 8 and reward 16, which gives inverse utility 0.5. This representation is convertible to the representation which assigns edge weight 1 to 2 as 1 and 2 to 3 as -0.5. }
\label{fig::transfer}
\end{figure}

It is necessary to have a polynomial-time algorithm to find the optimal utility path. Maximizing utility is equivalent to minimizing inverse utility, the ratio of risk to reward, and this resembles the shortest path problem. As shown in Fig. \ref{fig::transfer}, the reward risk representation as utility could be easily converted to edge weight representation. It is desirable that the optimal utility path could be found using traditional shortest path algorithms after converting the risk reward representation into edge weights. However, Dijkstra'a algorithm fails due to its positive weights assumption (negative edge weight from vertex 2 to 3 in Fig. \ref{fig::transfer}). Even with Bellman-Ford algorithm which can deal with negative weights, the existence of negative cycle makes it not applicable to our problem: it is easy to have a risk representation that makes cycling between vertex 2 and 3 collecting good rewards while facing only small amount of risk. Therefore the optimal utility path must be enforced to be simple, because otherwise the optimal solution is not well-defined: the agent will just looping around the negative cycle forever to infinitely decrease the optimal inverse utility. 

Given the easy conversion between our utility and edge weight representation (Fig. \ref{fig::transfer}), it could be shown that our risk-aware reward-maximizing problem is reducible from shortest simple path problem with negative cycle in the graph, which is further reducible from the longest path problem. It is well-known that longest path problem is NP-hard, thus our risk-aware reward-maximizing problem is NP-hard. 

Therefore in order to be computationally efficient, we need an approximate algorithm that can give suboptimal solution within a reasonable amount of time. 

\subsection{Approximate Algorithm}
The approximate algorithm is divided into two stages: upper stage plans minimum-risk paths from start to every other vertex using a search algorithm similar to Dijkstra's approach.  Lower stage planner selects the maximum utility path from the ensemble of minimum-risk paths provided by the upper stage planner. As shown in Fig. \ref{fig::approximate}, the final suboptimal path is found as the minimum-risk path to the same goal location as found by the exact algorithm, but the optimal path (Fig. \ref{fig::exact}) is neglected by the approximate algorithm. Fig. \ref{fig::both_algos} indicates that it is actually worth the extra risk caused by the extra tortuosity to collect more rewards along the path. But the approximate algorithm only looks at the ensemble of minimum-risk paths. 

\subsubsection{Upper Level Risk-Aware Planner}
\begin{figure}[]
\centering
\subfloat[Dynamical]{\includegraphics[width=0.5\columnwidth]{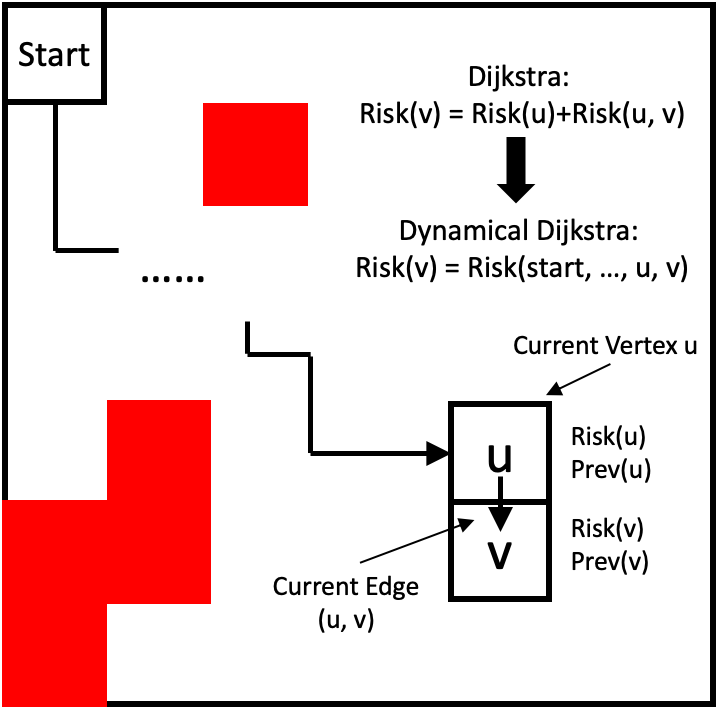}%
\label{fig::dynamical}}
\subfloat[Directional]{\includegraphics[width=0.5\columnwidth]{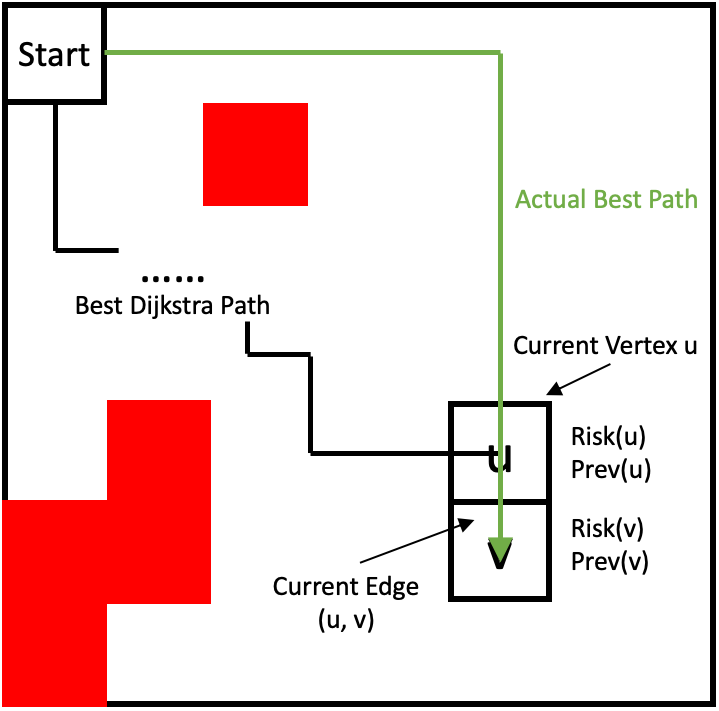}%
\label{fig::directional}}
\caption{Dynamical and Directional Components in Addition to Dijkstra's: path risk needs to be evaluated dynamically and non-optimal substructure requires minimum-risk path to be directional. }
\label{fig::DDD}
\end{figure}

The upper stage planner employs a similar approach to Dijkstra's, but with two major differences: dynamical and directional (Fig. \ref{fig::DDD}). The path-dependent explicit risk representation requires path risk to be evaluated based on the entire path, instead of only summation of current risk at $u$ and extra risk from $u$ to $v$ from regular Dijkstra's (Fig. \ref{fig::dynamical}). Being directional is mainly due to non-optimal substructure. Minimum-risk path to $v$ via $u$ may not include minimum-risk path to $u$. The risk to each vertex will depend on which direction the planner takes to get to the vertex and it will further non-additively affect future risk evaluation. It still requires the assumption of non-decreasing risk along the search, which is apparently true based on the explicit risk representation in Sec. \ref{sec::explicit_risk_representation}.

The graph is defined similarly as in exact algorithm by $\mathcal{G} = (\mathcal{V}, \mathcal{E})$. To suit the directional needs, each vertex is further decomposed by $v_i = (D_i^{(1)}, D_i^{(2)}, ..., D_i^{(c)})$, where $D_i^{(j)}$ represents the direction from which $v_i$ is reached. The total number $c$ is the connectivity of $v_i$, as the number of incoming edges reaching $v_i$. For each direction reaching $v_i$, $D_i^{(j)}$ is defined as $D_i^{(j)} = (r_i^{(j)}, PD_i^{(j)})$, where $r_i^{(j)}$ is the minimum risk of reaching $v_i$ from direction $D_i^{(j)}$, starting from start vertex $v_{start}$. $PD_i^{(j)}$ is the previous direction of reaching the previous vertex, in other words, previous direction of one more step ahead. All the directions of all vertices $D_i^{(j)}$ compose the superset of all directions $\mathcal{D} = \{D_i^{(j)}|i=1, 2, ..., n\}$ and $j$ is a variable for different vertices depending on how many directions (edges) are leading to the vertex. The algorithm is shown in Alg. \ref{alg::risk-aware}: instead of closing each vertex as in Dijkstra's, we close each direction (line 5 and 16) of each vertex (directional). The risk is evaluated based on the entire path (line 7 - 9) with the $backtrack$ function (dynamical). Given the current directional component, it is trivial to backtrack the previous vertex. And the one before is saved in $PD_i^{(j)}$ and $backtrack$ can easily find the history path leading to $V_{start}$.

\begin{algorithm}[]
 \caption{Risk-Aware Path Planner}
 \begin{algorithmic}[1]
 \renewcommand{\algorithmicrequire}{\textbf{Input:}}
 \renewcommand{\algorithmicensure}{\textbf{Output:}}
 \REQUIRE $\mathcal{G}$, $v_{start}$
 \ENSURE  Risk-Aware paths to all vertices other than $v_{start}$
  \STATE $\forall D_i^{(j)} \in \mathcal{D}$ set $r_i^{(j)} \leftarrow \infty$ and $PD_i^{(j)} \leftarrow NULL$
  \STATE For $v_{start}$, set $r_{start}^{(j)}\leftarrow 0$ in all $D_{start}^{(j)}$
  \STATE Initialize visited set to $\mathcal{R}\leftarrow\{\}$
  \WHILE {$\mathcal{R}\neq\mathcal{D}$}
  	\STATE pick vertex $v_u$ with smallest $r_u^{(i)}$ where $D_u^{(i)}\notin\mathcal{R}$
	\FOR {each edge $(v_u, v_v)\in\mathcal{E}$}
		\STATE $path_u^{(i)}\leftarrow backtrack(D_u^{(i)}$)
		\STATE $path_v(i) \leftarrow path_u^{(i)} \cup \{v_v\}$
		\STATE $path\_risk_v(i)\leftarrow evaluate(path_v(i))$
		\STATE $current\_min\_risk\leftarrow v_v.D_v^{(j)}.r_v^{(j)}$, where $D_v^{(j)}$ corresponds to reaching $v_v$ from $v_u$
		\IF {$path\_risk_v(i)<current\_min\_risk$}
			\STATE $v_v.D_v^{(j)}.r_v^{(j)}\leftarrow path\_risk_v(i)$
			\STATE $v_v.D_v^{(j)}.PD_v^{(j)} \leftarrow D_u^{(i)}$
		\ENDIF
	\ENDFOR
	\STATE $\mathcal{R}\leftarrow \mathcal{R} \cup \{D_u^{(i)}\}$
  \ENDWHILE 
  \FOR {each $v_i \in \mathcal{V}$}
  	\STATE pick $D_i^{(j)}$ with the smallest $r_i^{(j)}$
	\STATE $risk_i \leftarrow r_i^{(j)}$
	\STATE $path_i \leftarrow backtrack(D_i^{(j)})$
  \ENDFOR
 \RETURN all $path_i$ with $risk_i$
 \end{algorithmic}
 \label{alg::risk-aware}
 \end{algorithm}

\subsubsection{Lower Level Reward-Maximizing Planner}
The lower level planner only considers the ensemble of minimum-risk paths provided by the upper level planner and chooses the maximum utility path (Alg. \ref{alg::lower_stage_planner}). It is also necessary to compare with the utility of staying at $V_{start}$ given the possibility that the agent starts at a good viewpoint and it does not worth to take the risk to go anywhere else. 

\begin{algorithm}[]
 \caption{Maximum Reward Planner}
 \begin{algorithmic}[1]
 \renewcommand{\algorithmicrequire}{\textbf{Input:}}
 \renewcommand{\algorithmicensure}{\textbf{Output:}}
 \REQUIRE \textit{ensemble of minimum-risk paths}, \textit{reward map}, $\gamma$
 \ENSURE  sub-optimal utility path 
  \FOR {each \textit{path} in \textit{ensemble}}
	\STATE reward[\textit{path}] $\leftarrow$compute\_overall\_reward(\textit{path}, $\gamma$)
	\STATE risk[\textit{path}] $\leftarrow$ evaluate(\textit{path})
	\STATE utility[\textit{path}] $\leftarrow$ reward[\textit{path}] / risk[\textit{path}]
  \ENDFOR
  \STATE Compute utility of staying at \textit{start} as a unit \textit{path}
   \RETURN \textit{path} with maximum utility value
 \end{algorithmic}
 \label{alg::lower_stage_planner}
 \end{algorithm}

\begin{figure}[]
\centering
\includegraphics[width=0.6\columnwidth]{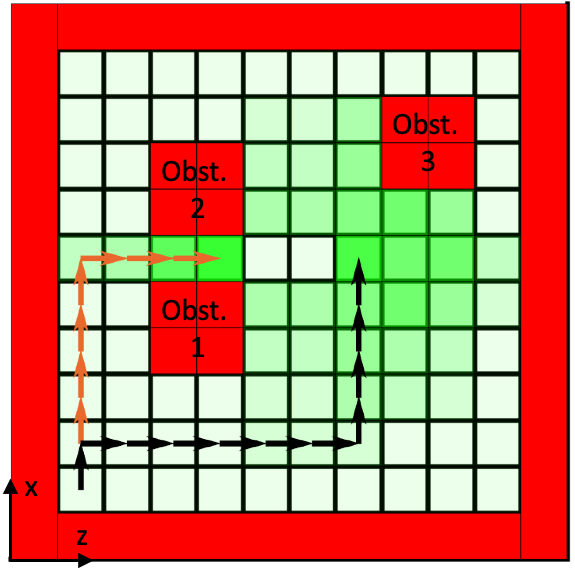}
\caption{Approximate Algorithm Solution: to observe the action taking place in the middle two white cells, the planner finds a path which minimizes both states and path risk and collects good rewards along the entire path. The orange path only aims at the best rewarding state but faces large risk. }
\label{fig::appriximate_example}
\end{figure}

One example solution of the approximate algorithm is shown in Fig. \ref{fig::appriximate_example}. 
The suboptimal path found by the approximate algorithm minimizes both states and path risk by going through wide open spaces between obstacles and making as few turns as possible, respectively. The reward collected along the entire path is maximized simultaneously. The best viewpoints, shown in green between the two obstacles, do not worth to go to due to the risk of going through tight spaces. Although the example is shown in 2-D, this algorithm works in any dimensions with any vertex connectivity. 

\section{Physical Demonstration}
\label{sec::experiments}
\begin{figure}[]
\centering
\includegraphics[width=1\columnwidth]{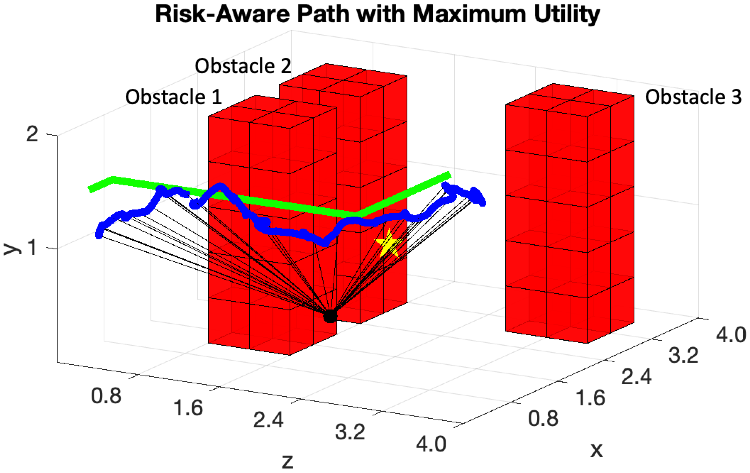}
\caption{Risk-Aware Path with Maximum Utility Value Executed on a Physical Tethered UAV: red voxels represent obstacles. Yellow star is the visual assistance PoI. The planned path is shown in green while the physically executed path in blue. }
\label{fig::experiment}
\end{figure}

\begin{figure}[]
\centering
\includegraphics[width=1\columnwidth]{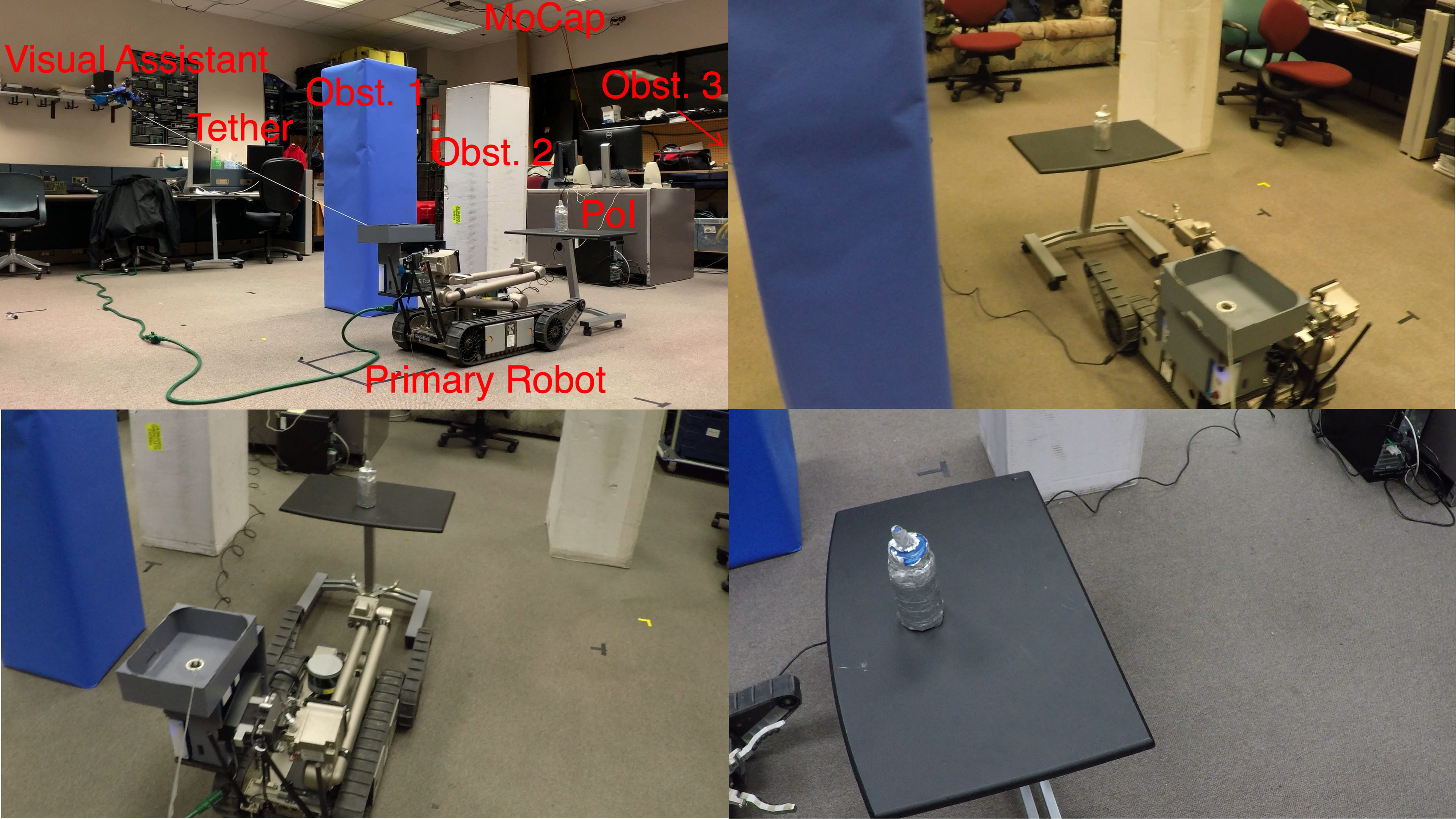}
\caption{Upper Left: Third Person (External) View of the Physical Demonstration. Upper Right, Lower Left and Lower Right: Accumulated Rewards in Terms of Visual Assistance Video Feed (Snapshots) along the Entire Risk-Aware Maximum-Utility Path in Chronological Order.}
\label{fig::all_snaps}
\end{figure}

In this section path shown in Fig. \ref{fig::appriximate_example} is implemented on a tethered aerial visual assistant, Fotokite Pro, using the low level motion primitives described in \cite{xiao2018motion, xiao2018indoor}. The physical demonstration aims at showing the proposed risk-aware reward-maximizing planner being used on real robot, with its actually encountered motion risk in physical environments (Fig. \ref{fig::experiment}) and real-world collected reward in terms of visual assistance quality (Fig. \ref{fig::all_snaps}). 

Our physical demonstration is conducted in a motion capture studio to ground-truth the visual assistant's actual motion. The studio is equipped with 6 OptiTrack Flex 13 cameras running at 120Hz. The 1280$\times$1024 high resolution cameras with a 56\degree~Field of View provide less than 0.5mm positional error and cover the whole 4$\times$4$\times$2m space. Although the original planner is shown in 2-D for easy illustration, the physical demonstration is conducted in 3-D space, with the same 3 obstacles distributed in the map. The mission for the tele-operated ground robot is to pick up a sensor in front with visual assistance from the tethered UAV. The visual Point of Interest (PoI) is therefore defined as the sensor, shown as the yellow star in Fig. \ref{fig::experiment}.

As shown in Fig. \ref{fig::appriximate_example}, the most rewarding state (best viewpoint) is to the left of the PoI (from ground robot's point of view). A traditional planner would plan a path leading to the optimal viewpoint (shown in orange). However, considering the fact that the best viewpoint locates between two obstacles and the path leading to it goes through the narrow passage between one obstacle and the map boundary (also treated as obstacle) and contains tether contact, it does not worth to take the risk. Our risk-aware reward-maximizing planner, on the other hand, could balance the trade-off between reward and risk. The approximate algorithm compares the utility value of the minimum-risk path leading to the optimal viewpoint with other candidate paths, and chooses the one with optimal utility among all minimum-risk paths. The planned path (green) and actual path (blue) in Fig. \ref{fig::experiment} maintain a maximum distance to closest obstacle and also a good visibility value and therefore a low states risk along the way, while making only two turns with zero tether contact to minimize path risk. 

\section{CONCLUSION}
\label{sec::conclusion}
This paper formally formulates a risk-aware reward-maximizing problem motivated by an autonomous visual assistance scenario in unstructured or confined environments. A new explicit motion risk representation framework is proposed, extending current risk evaluation's state-only dependency to a more comprehensive function of entire path. Based on this new risk representation, this work also presents a risk-aware path planner which maximizes accumulated reward simultaneously to balance the trade-off between mission reward and motion risk. The well-definedness and NP-hardness of this problem are proved, and then a two-stage approximate algorithm is provided. The high-level risk-aware planner adds dynamical and directional components into regular Dijkstra's algorithm, making the new algorithm suitable for problems with costs that need to be dynamically evaluated and without substructure optimality in general. We have demonstrated the planned sub-optimal utility path on a physical tethered UAV, which locomotes in a risk-averse manner while also collecting real-world reward as visual assistance viewpoint quality. The physical motion risk and mission reward of the path are presented and discussed. 


\section*{ACKNOWLEDGMENT}
This work is supported by NSF 1637955, NRI: A Collaborative Visual Assistant for Robot Operations in Unstructured or Confined Environments. 

\bibliographystyle{IEEEtran}
\bibliography{IEEEabrv,references}

\end{document}